\documentclass[times, 10pt,twocolumn]{article} 
\usepackage{latex8}
\usepackage{amsmath,graphicx}
\usepackage[hyphens]{url}
\usepackage{amsmath}
\usepackage{paralist} 
\usepackage{booktabs}
\usepackage[dvipsnames]{xcolor}
\usepackage{microtype}
\usepackage{ragged2e}

\PassOptionsToPackage{hyphens}{url}
\usepackage{caption}
\usepackage{subcaption}
\usepackage{enumitem}
\RequirePackage{xcolor}
\usepackage{multirow}
\usepackage[hyphens]{url}
\usepackage[pagebackref=false,breaklinks=true,colorlinks,bookmarks=false]{hyperref}

\usepackage{color}

\usepackage{latex8}
\usepackage{times}

\begin{document}

\title{Open-Eye: An Open Platform to Study Human Performance on Identifying AI-Synthesized Faces}

\author{Hui Guo$^{1}$, \ Shu Hu$^{2}$, \ Xin Wang$^{2}$, \   Ming-Ching Chang$^{1}$, \  Siwei Lyu$^2$ \\
$^{1}$University at Albany, SUNY, USA \ \ \  $^{2}$University at Buffalo, SUNY, USA \\
{\tt\small \{hguo, mchang2\}@albany.edu}, \ \ \  {\tt\small \{shuhu, xwang264, siweilyu\}@buffalo.edu}
}



\maketitle
\thispagestyle{empty}

\begin{abstract}
AI-synthesized faces are visually challenging to discern from real ones. They have been used as profile images for fake social media accounts, which leads to high negative social impacts. 
Although progress has been made in developing automatic methods to detect AI-synthesized faces, there is no open platform to study the human performance of AI-synthesized faces detection. 
In this work, we develop an online platform called Open-eye to study the human performance of AI-synthesized faces detection. We describe the design and workflow of the Open-eye in this paper.
\end{abstract}

\vspace{-0.3cm}
\Section{Introduction}
\vspace{-0.2cm}

\label{sec:intro}

The rapid development of Artificial intelligence (AI) technologies, called Deep Fakes, has made it possible to synthesize highly realistic images, audio, and video that are difficult to discern from real ones. In particular, AI-synthesized faces have been misused for malicious purposes. Recent years have seen an increasing number of reports that AI-synthesized faces were used as profile images on fake social media accounts, which generates negative social impacts~\cite{theverge,cnn1,cnn2,reuters}.

Generative adversarial networks (GANs) are the most popular technologies for synthesizing content and especially the human face images~\cite{karras2017progressive,karras2019style,karras2020analyzing,nightingale2021synthetic}. The pernicious impact of these synthesized faces has led to the development of methods aiming to distinguish GAN-generated images from real ones. 
Many of those methods are based on deep neural network (DNN) models due to their high detection accuracy~\cite{marra2019gans,wang2020cnn,guo2021robust}. 
Although they have achieved high accuracy and the models are end-to-end, the DNN-based methods suffer from several limitations. The models lack interpretability to the detection results, and they usually have a low capability to generalize across different synthesis methods \cite{hu2020learning, hu2021sum}.

\begin{figure*}[t]
\centerline{
  \includegraphics[width=0.98\textwidth]{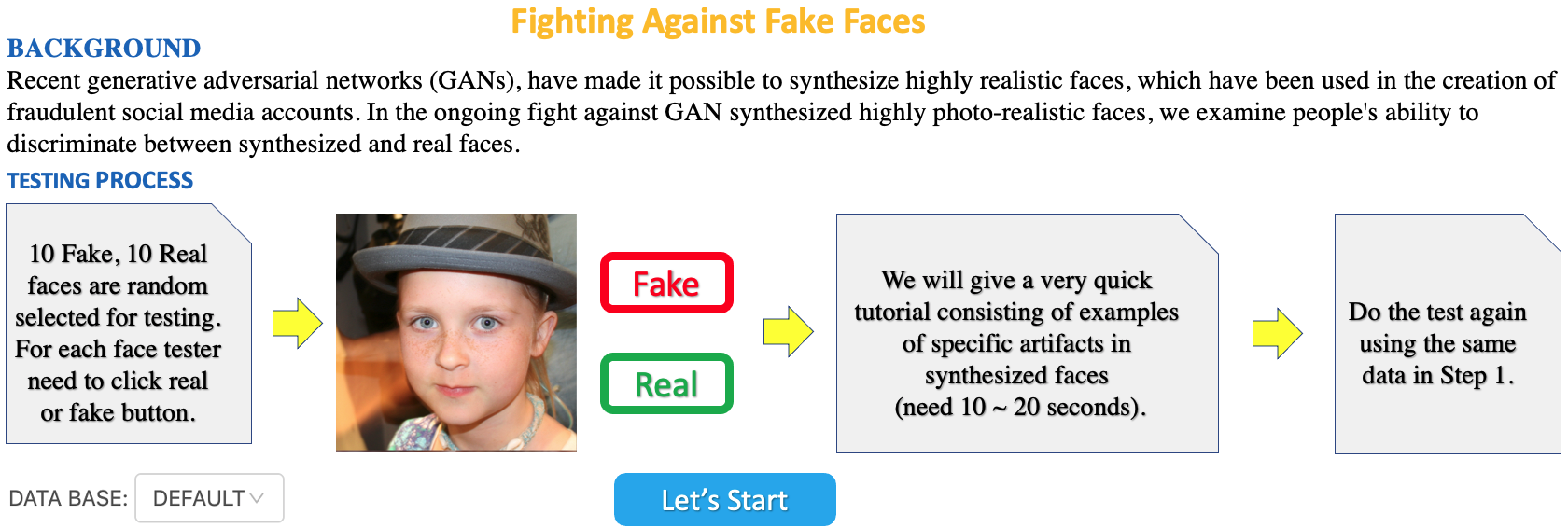}
  \vspace{-0.1cm}
}
\caption{\em The workflow of Open-eye. The participants test 20 face images, including the same amount of real and fake. A quick tutorial is demonstrated to participants to learn how to recognize the specific artifacts in the eye. The same test will be used again to show the human performance of the AI-synthesized faces detection. 
}
\vspace{-1em}
\label{fig_pipeline}
\end{figure*}  

As the existing methods are either less efficient or not accurate enough to handle the torrent of daily uploads of the public content \cite{wang2022gan}, the users must have the ability to recognize the fake faces from the real ones. Recently, the studies investigating the human performance of AI-synthesized faces detection have been conducted \cite{lago2021more,nightingale2021synthetic}. For example, \cite{nightingale2021synthetic} examined people's ability to discriminate GAN faces from real faces.
Specifically, 400 StyleGAN2 faces and 400 real faces from the FFHQ dataset are selected with large diversity across the genders, ages, races, {\em etc.}, and two sets of experiments are conducted. 
In the first set of experiments, 315 participants were shown a few examples of GAN faces and real faces, and around $ 50\%$ of accuracy was obtained. In the second set of experiments, 170 new participants were given a tutorial consisting of examples of specific artifacts in the GAN faces. 
Participants were also given feedback afterward. However, it was found that such training and feedback only improve a little bit of average accuracy. Therefore, this work concluded that the StyleGAN2 faces are realistic enough to fool naive and trained human observers. 
However, this study provides no information on what synthesis artifacts are provided for participant training. There is no open platform for public use to test.


To overcome the above limitations, we propose an open platform, Open-eye. The Open-eye consists of several steps (See Figure \ref{fig_pipeline}). In Stage 1, the participants are tested with real and AI-synthesized faces. 
In Stage 2, the participants are trained with the artifacts to identify AI-synthesized faces reliably. Because we believe there is still space to improve human capability in discerning AI-synthesized faces if sufficient hints are provided, such as philosophical cues ({\em e.g.} pupil shapes \cite{guo2021eyes}). In Stage 3, the participants will do the test again using the same data in Step 1.
Finally, the testing performance of both stages will show for comparison.
Our Open-eye platform is available at {\footnotesize \url{http://zinc.cse.buffalo.edu/ubmdfl/human-performance-on-gan-face/#/}}.


The main contributions of this work are two-fold:
\vspace{-0.2cm}
\begin{itemize}[leftmargin=16pt] \itemsep -.2em

\item We are the first to propose an open platform to study whether human participants can distinguish state-of-the-art AI-synthesized faces from real faces visually.

\item The proposed platform is flexible to incorporate any AI-synthesized faces and provide quick training to the participants to recognize the fake faces. An example shows that the platform is simple, effective, and efficient for participants.

\end{itemize}


\Section{Background}

\label{sec:related}
\vspace{-0.2cm}

\textbf{AI-synthesized faces.} One of the most popular AI-synthesized faces techniques is based on GAN models. A GAN model includes two neural networks (generator and discriminator) trained in tandem. The generator takes random noises as input and can effectively encode rich semantic information in the intermediate features and latent space for high-quality face image generation. The discriminator aims to distinguish synthesized images from the real ones. Generator and discriminator compete with each other during the training.  
A series of GAN models (\textit{e.g.}, PGGAN \cite{karras2017progressive}, BigGAN \cite{brock2018large}, StyleGAN \cite{karras2019style}, StyleGAN2 \cite{karras2020analyzing}, StyleGAN3 \cite{karras2021alias}) have been developed and demonstrated superior capacity in generating or synthesizing realistic human faces.  In some early works such as \cite{yang2019exposinggan}, they find that faces generated by the early StyleGAN model~\cite{karras2019style} \cite{karras2019style} have considerable artifacts such as fingerprints~\cite{marra2019gans, yu2019attributing}, inconsistent iris colors~\cite{li2018detection, mccloskey2018detecting}, {\em etc.}. However, just one year later, StyleGAN2 is proposed by Karras et al. in \cite{karras2020analyzing} and it has greatly improved the visual quality and pixel resolution, with largely-reduced or undetectable artifacts in the generated faces. 

{\bf AI-synthesized faces detection.}
With the development of the GAN models for face generation/synthesis, methods for distinguishing GAN-generated faces have progressed accordingly. Most of these methods are Deep Learning based~\cite{marra2019incremental,hulzebosch2020detecting,wang2020cnn,goebel2020detection,liu2020global}. 
Notably, several methods exploit the physiological cues (which suggest inconsistency in the physical world) to distinguish GAN-generated faces from the real ones~\cite{matern2019exploiting}. 
In \cite{yang2019exposinggan}, GAN-generated faces are identified by analyzing the distributions of the facial landmarks. 
The work of \cite{hu2021exposing} analyzes the light source directions from the perspective distortion of the locations of the specular highlights of the two eyes. 
Such physiological/physical-based methods come with intuitive interpretations and are more robust to adversarial attacks~\cite{verdoliva2020media, hu2021tkml}.

\begin{figure*}[t]
\centerline{
  \includegraphics[width=0.78\linewidth]{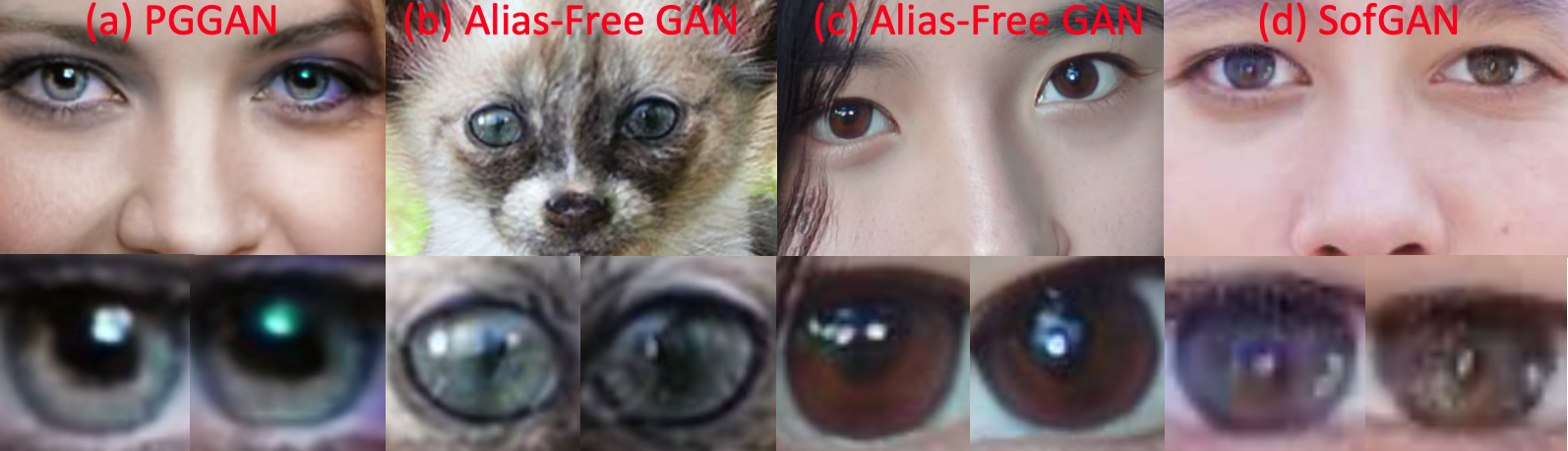}
  \vspace{-2mm}
}
\caption{\em Examples of GAN-synthesized faces additional to StyleGAN and StyleGAN2. The images are from their original papers \textbf{(a)} PGGAN \cite{karras2017progressive}, \textbf{(b,c)} Alias-Free GAN (StyleGAN3) \cite{karras2021alias}, \textbf{(d)} SofGAN \cite{chen2021sofgan}. Observe in the zoomed-in view that the pupils appear in irregular, inconsistent shapes, which tell them apart from real faces. }
\vspace{-0.2cm}
\label{fig:other:GANs}
\end{figure*}  


\begin{figure*}[t]
\centerline{
  \includegraphics[width=\textwidth]{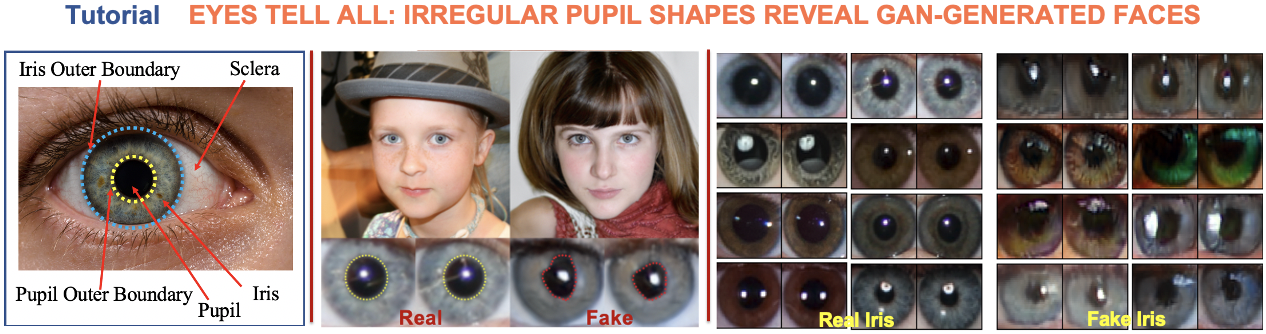}
  \vspace{-0.1cm}
}
\caption{\em Tutorial (Stage 2). \textbf{Left:} Anatomy of a human eye: the iris and pupil are at the center surrounded by the sclera. \textbf{Middle:} Examples of pupils of real human (left) and GAN-generated (right). Note that the pupils for the real eyes are in circular or elliptical shapes (yellow), while those for the GAN-generated pupils are much irregular (red). For GAN-generated faces, the shapes of the pupils are very different from each other when zoomed-in. \textbf{Right:} More pair of pupil examples from real (left) human faces and GAN-generated (right) faces. As we mentioned before, the irregular pupil shape is a good sign for the human to expose the GAN-generated face visually, we can easily see that the shapes of GAN-generated pupils are very irregular, and the shapes of both pupils are very different in the same GAN-generated face image. In practice, people can zoom a face image large enough and then check the pupil shapes to find whether the face is real or not easily. 
}

\vspace{-1em}
\label{fig:pipeline}
\end{figure*}  

\section{Platform Design}
\vspace{-0.2cm}


This section describes the design of the Open-eye platform. Our platform is composed of three stages: 

In \textit{Stage 1}, the participants are tested with real and AI-synthesized faces from a given dataset. need to select the dataset 
In \textit{Stage 2}, the participants are trained with the artifacts to identify AI-synthesized faces reliably. Our participants' training method is motivated by observing that GAN-generated faces exhibit a common artifact. 
For example, Iris and pupil Analysis is a critical task in biometric identification that has been studied well. The pupils appear with irregular shapes or boundaries, other than a smooth circle or ellipse. 
This artifact is universal for all known GAN models (at least for now, {\em e.g.} PGGAN~\cite{karras2017progressive}, Alias-Free GAN~\cite{karras2021alias}, and SofGAN~\cite{chen2021sofgan}), as shown in Figure \ref{fig:other:GANs}. 
In \textit{Stage 3}, the participants will do the test again using the same data in Step 1. The overview of the platform workflow is illustrated in Figure \ref{fig_pipeline}.

In the next section, we describe the use of the Open-eye platform with the above example.

\vspace{-1mm}
\section{An Example}
\vspace{-2mm}

In this section, we use irregular pupil shapes to reveal
GAN-generated faces \cite{guo2021eyes} as a tutorial example to further explain the platform. In general, we can incorporate more methods and datatype into our platform. In general, we can incorporate more methods and datatype in our platform.   

\textbf{Datasets.} 
We use the real human faces from the Flickr-Faces-HQ (FFHQ) dataset~\cite{karras2019style}. 
Since StyleGAN2~\cite{karras2020analyzing}~\footnote{\url{http://thispersondoesnotexist.com}} is currently the state-of-the-art GAN face generation model with the best synthesis quality, we collect GAN-generated faces from it. We only use images where the eyes and pupils can be successfully extracted. 

In \textit{Stage 1}, 20 images are shown to the participants. Note that 20 is only a predefined number. We can adjust it as the demand need. 10 real images are randomly selected from the Flickr-Faces-HQ (FFHQ) dataset~\cite{karras2019style}, and 10 fake images are randomly selected from StyleGAN2~\cite{karras2020analyzing}. The participants are requested to click the real or fake button for each image based on their cognition. After the test, the system will output the human performance, including accuracy, precision, recall, and F-score.

In \textit{Stage 2}, three courses are provided to the participants that can be learned to recognize the fake images from real ones, including introducing human eye anatomy, comparing pupils from real human faces and AI-synthesized faces, and presenting more iris examples to show the difference between the real and fake pupils (see Fig. \ref{fig:pipeline} for more details). Specifically, the participants will be taught to identify the pupil outer boundary, iris outer boundary, sclera, iris, and pupil in an eye. Then, the participants will learn the difference between real and fake pupils, \textit{e.g.}, fake pupils may contain unclear boundaries, and the shape stretched up or toward the width. This stage mainly brings an awareness of how fake and real pupils are distinguished in shapes.

In \textit{Stage 3}, the participants will do the test again using the same 20 images from Stage 1. We will also output the new human performance. By comparing two results (in Stage 1 and Stage 3) from the same data set, we can get a cognition of whether human participants can distinguish state-of-the-art AI-synthesized faces from real faces visually after learning from the tutorial.

The proposed pupil shape-based tutorial contains several limitations. Since the method is based on the simple assumption of pupil shape regularity, false positives may occur when the pupil shapes are non-elliptical in the real faces. 
This may happen for infected eyes with certain diseases. Also, poor imaging conditions, including lighting variations, largely skewed views, and occlusions, can cause errors in pupil segmentation or thresholding errors. To make our system more useful and educational, we can add more tutorials according to existing research. For example, \cite{hu2021exposing} uses the inconsistency of the corneal specular highlights between the two synthesized eyes to identify GAN-generated faces, and \cite{matern2019exploiting} discerns the GAN-generated faces by different iris \cite{wang2021nir} colors of the left and right eye.



\vspace{-0.2cm}
\section{Conclusion}
\vspace{-0.3cm}


In this work, we describe an open platform, known as Open-eye, for investigating the human performance of AI face detection. This platform provides interfaces for training the participant that may further improve forensic detection effectiveness. 
For future works, we will continue to integrate more AI faces into the platform that can further expand the impact in addressing issues in social media forensics. And investigate the human performance with this open platform.





\bibliographystyle{latex8}
\bibliography{latex8}

\end{document}